# A Modified Drake Equation For Assessing Adversarial Risk To Machine Learning Models

Josh Kalin[1,2], David Noever[2], and Matthew Ciolino[2], Gerry Dozier[1]

[1]Department of Computer Science and Software Engineering, Auburn University, Auburn, AL, USA
jzk0098@auburn.edu, doziergz@auburn.edu

[2]PeopleTec, Inc, Huntsville, AL, USA
David.Noever@peopletec.com, Matt.Ciolino@peopletec.com

## ABSTRACT

Machine learning models present a risk of adversarial attack when deployed in production. Quantifying the contributing factors and uncertainties using empirical measures could assist the industry with assessing the risk of downloading and deploying common model types. This work proposes modifying the traditional Drake Equation's formalism to estimate the number of potentially successful adversarial attacks on a deployed model. The Drake Equation is famously used for parameterizing uncertainties and it has been used in many research fields outside of its original intentions to estimate the number of radio-capable extra-terrestrial civilizations. While previous work has outlined methods for discovering vulnerabilities in public model architectures, the proposed equation seeks to provide a semi-quantitative benchmark for evaluating and estimating the potential risk factors for adversarial attacks.

## INTRODUCTION

This short note explores a simple version of a probabilistic equation for machine learning (ML), specifically to defend trained models from adversarial attacks. Probabilistic frameworks like the Drake Equation (predicting the number of alien civilizations [1-2]) offer a heuristic for traditional factor analysis, particularly helpful in the face of uncertainty. In that spirit, we adopt the basic formalism of its original population-based assessment by including both the leading factors and their corresponding loss fractions. The goal is less to provide a new ML risk model as much as to explore the practical factors needed to consider before fielding a new model [3]. Other work has detailed already much of the successes and failures in defending predictive models:
1) diversifying or augmenting training data [5]
2) ensembling and model voting [6]
3) obfuscating [7]
4) active learning [8]
5) protecting model descriptions or firewalling (e.g. nation-state security) [9]

All these strategies (plus many others not included here [3]) should appear in one or more proposed terms for this modified Drake Equation.

## MODIFYING THE DRAKE EQUATION

The format of the note is first to present the main factors for model defense, followed by an explanation and examples to support each included factor. We explore the interpretation of each factor where possible with an illustrative case mined from the AI Incidents database [4]. It is worth noting that other than these early efforts [4] to catalog adversarial attacks, less research has previously attempted to count or quantify systematically the failures of a given machine learning model in the wild. For example, should an adversarial attack be scored based on its frequency, severity, or difficulty to patch once a vulnerability gets discovered? This paucity of data further motivates the development of a modified Drake Equation, principally as a heuristic framework for understanding contributing factors and assessing their uncertainties. The structure of the note isolates each factor in reference to its more familiar population-based input, so for instance, the time that an attacker might probe a model's vulnerabilities maps to the original Drake Equation's reference to the time that a radio-aware civilization might broadcast its identity. An appealing aspect of the original format stems from its hierarchical factors from large to small fractional contributions as they change over time. One ultimately wants to understand the dependencies while solving for the machine learning model's attack surface, as measured by N, the number of successful adversarial attacks.

To defend a machine learning model, the number of successful adversarial attacks, N, is proportional to the model's size, R, as measured by its popularity (e.g. YOLO), sponsoring enterprise size (e.g. Microsoft, Google, Facebook), or monoculture of adoption (e.g. convolutional neural networks). The proposed modifications to the Drake Equation are described below:

$$N = R * f_p * n_e * f_l * f_i * f_c * L$$

$N = $ the number of successful adversarial attacks
$R = $ average enterprise size
$f_p = $ fraction of models published, named, open sourced or fielded in the wild
$n_e = $ average number of **e**ngineered parameters (memory, billions of parameters)
$f_l = $ fraction of **l**earning ratio, as training/test data or active hybrid feedback
$f_i = $ fraction of **i**nput supervisory and quality control steps
$f_c = $ fraction of **c**ompleted queries that return detectable or logged answers
$L = $ **l**ength of time that attackers can query without consequences or timeouts

### R - Average Enterprise Size
In the original Drake Equation, this factor traditionally relates to a rate of new star formation. We generalize the rate of new ML models created, R, by an aggregate of overall enterprise size. This approach mirrors the literature on monoculture in computer operating systems (e.g. MS Windows) as a primary indicator to motivate cyber-attacks. The corresponding figure in defending ML models derives from a similar feature, namely that attacking large enterprise models like Google's Perspective API and OpenAI's Generative Pretrained Transformer (GPT-3) is more likely than probing or disabling a smaller, private, or novelty ML model.

One can hypothesize that the community's attraction to leader boards [4] and state-of-the-art (SOTA) competitions further drives the ML community to more singular ecosystems that may prove more difficult to defend from adversaries than a diversified one. As a figure of merit when describing the cyber-risks for a monopoly in operating systems [10], the entire ecosystem may become unstable when the market share and global adoption reach 43% and more directed attacks garner hacker's attention. One ML-specific metric of whether a given type of neural network dominates its ecosystem can be approximated by search trend monitors. For example, by using Google Trends [11], the current popularity of three core approaches to modeling the neural architecture itself shows that convolutional networks (CNN) capture 72% market share, compared to graph neural networks (25%) and capsule networks (2%). An attacker that knows the unique weaknesses of CNNs (such as their inability to grasp long-range spatial relations. complex hierarchies, and symmetries [12-13]) may profitably attack those specific design elements, particularly given their monopoly as deployed models.

### $f_p$ - fraction published, named, open-sourced, or fielded in the wild
In the original Drake Equation, this first factor in a hierarchical loss fraction stems from the number of stars with planets. In an adversarial attack, this factor similarly appears at the top of the hierarchy, namely how much is known about the model's origins. The literature spans model security from black-box (no knowledge) to white-box (full-knowledge), such that given a known or discoverable model structure, the attacker may also potentially know the weights and training dataset. This is most well-known in the release of GPT-2 versus GPT-3, where for some time the GPT-3 API was not available to researchers. When Open AI initially open-sourced its models, the company furthermore specifically withheld its larger one (1554M) to suppress the possibilities for abuse.

### $n_e$ - average number of engineered parameters
In the original Drake Equation, this second factor considers the number of planets capable of supporting life. In an adversarial attack, the relevant context would include model complexity, either as its memory, number of parameters, or layers of network architecture. A breakdown of computing $n_e$ for a CNN could be as simple as a baseline of the number of parameters or number of layers. For object detectors, the relevant complexity often arises from the way the model searches for its characteristic anchor sizes and bounding boxes, whether multi-stage like a region-proposal network (R-CNN) or single-stage frameworks like YOLO.

### $f_l$ - fraction of learning ratio
In the original Drake Equation, this third factor refers to planets that spawn life at some point. In an adversarial attack, this fraction includes losses for well-trained models that possess large and diverse data. Previous work has proposed using Akaike information criterion (AIC) and Bayesian information criterion (BIC) to evaluate model performance against dataset diversity and this style of metric may provide a baseline for this factor [14].

### $f_i$ - fraction of input supervisory guidance
In the original Drake Equation, this fourth factor addresses the rise of intelligent life from more primitive forms. In the machine learning context, this fraction includes the standard quality checks that separate a fieldable model from an experiment or lab bench demonstration. This factor corresponds to the breakpoint in many adversarial defenses, such that a prototype moves into production based on disciplined quality checks. Has the model seen out-of-vocabulary terms if a natural language processor? Is there a high fraction of augmented examples for each class? One traditional image approach augments training data with more diverse object types, usually including different lighting, viewing angles, or noise. Paleyes, et al. [15] describe 38 factors attacking 15 systems that contribute to a failed productization of an ML model. At any one of these steps, ranging from data-collection to performance monitoring, there exist adversarial attacks that can poison the entire process. Wang et al. [16] define in detail the adversarial attack to each of these systems.

### $f_c$ - fraction of completed queries that return detectable or logged answers
In the original Drake Equation, this fifth factor delineates the rise of technological capabilities such as radio transmission that travels at the speed of light and thus renders a distant galaxy observable. For adversarial attacks, this fraction defines the likelihood that an outside observer can understand the model type, its sensitivities, or its vulnerabilities. Particularly in the black-box approach where an attacker must launch a question-and-answer format to understand how the model works, this fraction restricts the obtainable universe of effective attacks. In experiments for text and image classifiers, Kalin, et. al [17] found that model architectures are easily discovered with strategic probing if the architecture is public. In this new equation, $f_c$ is related proportionally to $f_p$ factor.

### *L – Length of time that attackers can query without consequence or timeouts*
In the original Drake Equation, this final factor introduces the notion of time, particularly how long a civilization might survive its technology before self-destructing or its evolutionary time to propagate signals to an outside observer. Like the numerical count of accepted API requests ($f_c$), the length of time to automate or web-scrape the API with new queries offers a secondary line of defense not in space (count) but in time. Despite a more mature field, software engineering for APIs still suffers from vulnerable code being written into production systems [18].

## *MISSING BUT NOT FORGOTTEN*

This modification of the Drake Equation focuses on metrics that can be directly measured in a production environment. Missing elements in this heuristic might include additional pre-production factors for diversity, size, and quality of the input data, training lengths (epochs), and other historical elements that may or may not propagate usefully to the final model and its vulnerabilities. The collected metrics can then be used to refine the model performance against known benchmarks. For instance, common model types are easily discoverable via their input data and/or architecture [17].

## *AXIOMS*

### *Axiom 1: Architecture and Dataset Metrics are related*
The Learning Ratio, Parameters, and Guidance variables stem from the architectural design of the model. This equation is divided into two primary Adversarial fractional components: Architecture and Dataset. For teams to use the likelihood of successful adversarial attack assessment to improve their models, they will need to understand the contribution of architecture and dataset design to the overall adversarial risk. The first fraction defines the key parameters related to architecture and their overall contribution to adversarial risk:

$$Adversarial\ Fraction_{Architecture} = R * \frac{f_p * n_e * f_l}{N}$$

The second fraction defines the dataset metrics responsible for dataset contributions to adversarial risk:

$$Adversarial\ Fraction_{Dataset} = R * \frac{f_i * f_c * L}{N}$$

## *EXPERIMENTS*

This framework is designed to work on large and small works. In the following experiments, the focus is on baselining the effective ranges of the factors, showing sample risk factors for common model architectures, and understanding the relative effect each factor has on itself and the risk factor.

### *Experimental Design*
Each factor needs to be defined in terms of operating bounds to apply this new framework to current models. For the experiments, the following operating ranges for the variables were chosen to highlight the current capabilities that exist within the machine learning community today:
- **R - Average Enterprise Size**
    - Range [0, n authors]
    - Enterprise Size is computed as the Number of Authors as it can be difficult to find the actual number of employees in a particular organization
- *$f_p$ - fraction published, named, open-sourced, or fielded in the wild*

- Three values: Not Published 0.0, Published but not open source 0.5, Published and Open Source 1.0
- For example, GPT-3 is fielded in the wild but is not open source: 0.5
- $n_e$ - *average number of engineered parameters*
  - Stepped Range [0,1] based on Number of Model Parameters
- $f_l$ - *fraction of learning ratio*
  - Stepped Range [0,1] as a relative factor to State of the Art (SOTA) performance
  - For example, the first benchmark in the model category is 0.1 and SOTA is 1
- $f_i$ - *fraction of input supervisory guidance*
  - Range [0,1]
  - Is training data sufficiently large and diverse?
- $f_c$ - *fraction of completed queries that return detectable or logged answers*
  - Range [0,1]
  - Estimated High Query Rate on the model
- *L – Length of time that attackers can query without consequence or timeouts*
  - How long has the model been in public? Years [0, n]

As with the original formulation of the Drake Equation, each parameter represents an estimate of best guesses for factors in the wild. This modification to the Drake Equation will provide organizations the ability to benchmark, evaluate, and track the adversarial risk of their models in production. As a team observes the adversarial risk reduction on their model, there are factors within this equation that can directly be attributed to that reduced risk.

*Empirical Results*

Using the factor ranges described in the experiment design, six popular models were estimated as samples of how to apply this formulation. Figure 1 is sorted from top adversarial risk to lowest risk. In the example of 'MyModel', the model is not deployed and therefore does not contain adversarial risk from outside actors.

| Model | R | $F_p$ | $N_e$ | $F_l$ | $F_i$ | $F_c$ | L | $A_a$ | $A_d$ | N |
|---|---|---|---|---|---|---|---|---|---|---|
| T5 | 9 | 1 | 0.8 | 1 | 1 | 1 | 2 | 0.50 | 1.25 | 14.40 |
| VGG19 | 2 | 1 | 0.6 | 1 | 1 | 1 | 6 | 0.17 | 1.67 | 7.20 |
| GPT3 | 31 | 0.5 | 1 | 1 | 0.75 | 0.5 | 1 | 2.67 | 2.00 | 5.81 |
| BERT | 4 | 1 | 0.6 | 0.75 | 1 | 1 | 2 | 0.50 | 2.22 | 3.60 |
| FastText | 4 | 1 | 0.1 | 0.7 | 1 | 1 | 4 | 0.25 | 14.29 | 1.12 |
| MoibleNetV2 | 5 | 1 | 0.1 | 0.5 | 0.5 | 1 | 3 | 0.67 | 20.00 | 0.38 |
| MyModel | 1 | 0 | 0.2 | 0.75 | 0.2 | 0.05 | 1 | | | 0.00 |

*Figure 1: Summary of Models Explored with Modified Drake Equation. Six Popular model architectures are benchmarked along with a custom model based on MobileNetV2's design. The table is sorted by estimated Adversarial Risk N in the last column. $A_a/A_d$ represent the Adversarial Fraction for architecture and dataset respectively.*

When exploring this formulation, it's incredible to see that newer, larger architectures are less vulnerable than older models. This is on purpose though as older models will have more vulnerabilities appear since they have been in circulation longer. There are further improvements that could be made to these experiments – for instance, the exploration of architectures could be split into text and computer vision. Each category of model architectures can have its boundary conditions. For instance, transformers technologies like BERT and GPT have revolutionized NLP problems over the last few years. Their properties may warrant a deeper exploration of parameter dependencies.

*Correlation Analysis*

The next experiment in this work is to understand the dependency of each factor on adversarial risk. Building a correlation matrix using the assumptions above, Figure 2 shows the relative importance of each factor to itself, the other factors, and to adversarial risk.

| X-Correl | R | $F_p$ | $N_e$ | $F_l$ | $F_i$ | $F_c$ | L | N |
|---|---|---|---|---|---|---|---|---|
| R | 1.000 | -0.061 | 0.619 | 0.305 | 0.049 | -0.083 | -0.439 | 0.116 |
| $F_p$ | -0.061 | 1.000 | 0.034 | -0.063 | 0.788 | 1.000 | 0.606 | 0.406 |
| $N_e$ | 0.619 | 0.034 | 1.000 | 0.848 | 0.428 | 0.018 | -0.240 | 0.669 |
| $F_l$ | 0.305 | -0.063 | 0.848 | 1.000 | 0.439 | -0.073 | 0.038 | 0.735 |
| $F_i$ | 0.049 | 0.788 | 0.428 | 0.439 | 1.000 | 0.783 | 0.482 | 0.599 |
| $F_c$ | -0.083 | 1.000 | 0.018 | -0.073 | 0.783 | 1.000 | 0.611 | 0.404 |
| L | -0.439 | 0.606 | -0.240 | 0.038 | 0.482 | 0.611 | 1.000 | 0.145 |
| N | 0.116 | 0.406 | 0.669 | 0.735 | 0.599 | 0.404 | 0.145 | 1.000 |

*Figure 2: Cross-correlation of variables to the Modified Drake Equation including Adversarial Risk.*

Within Figure 2, there are a few surprising things that come out of the correlation analysis. Here are the key observations:
- The most correlated variable to adversarial risk is fraction of the learning ratio ($\rho(f_l, N) = 0.735$)
- The fraction of learning ratio is highly correlated to the number of parameters ($\rho(f_l, N_e) = 0.848$)
- The fraction of input supervisory guidance is correlated to fraction published ($\rho(f_i, f_p) = 0.788$)
- The fraction completed queries is highly correlated to fraction published ($\rho(f_c, f_p) = 1.000$)

Intuitively, the fraction of learning ratio being most correlated to adversarial risk represents that the most popular models have the most people trying to attack them. The goal is to track and reduce the adversarial risk to a model and this framework provides a starting benchmark.

## SUMMARY AND FUTURE WORK

This work supports an established heuristic framework in analogy to the traditional Drake Equation. This simple formalism amounts to a summary of relevant factors. The basic equation has been modified elsewhere for detecting biosignatures in planet-hunting (Seager equation [19]), sociology (best choice problem [20]), infection risks [21], AI singularity [22], social justice [23], and other diverse probabilistic assessments [24]. Ultimately, its main purpose follows from assessing the multiple uncertainties that may vary by several orders of magnitude. For example, as ML builders consider whether to privatize or to open-source their models, they may intuitively favor one course over another given a perceived risk for model compromise. Is it true in practice that privatizing a model lowers the risk, or does it increase the attack surface because the model never gets hardened by peers? One would like to provide a framework for these important decisions and assist the ML community to identify the data needed for sensitivity analysis and the evaluation of consequences.

The biggest challenge in finding novel utility for this framework shares much in common with Drake's original notion. How to quantify each factor? What if the factors show strong correlations? How do the factors change with time, particularly if both the builders and attackers modify their behavior? What are the appropriate units to assess ML risks, either as the number or severity of adversarial attacks? One informative output that previous technical papers often ignore in assessing model risk is the scale of the overall ecosystem (R). In the literature for cybersecurity, for example, the monoculture aspect for operating systems has proven most predictive of the next generation's attacks. In this view, the SOTA leader boards [4] might benefit from encouraging a more diverse model ecosystem, such that niche YOLO attacks cannot propagate throughout the whole ML community and its applications, particularly when a few fractional percentage improvements separate the field into universal adoption strategies. Future work should highlight the data sources for evaluating each factor. For instance, the publications dataset from Cornell's arXiv supports extensive topic analysis for extracting the popularity of ML models, their relevant attack methods, and promising defensive styles [26]. Classification methods for attack types [27] may also guide the practical counting or scoring for the universe of adversarial ML threats.

## ACKNOWLEDGMENTS

The authors would like to thank the PeopleTec Technical Fellows program for its encouragement and project assistance. The views and conclusions contained in this paper are those of the authors and should not be interpreted as representing any funding agencies.

## AUTHORS

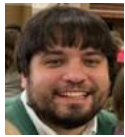
Josh Kalin is a physicist and data scientist focused on the intersections of robotics, data science, and machine learning. Josh holds degrees in Physics, Mechanical Engineering, and Computer Science.

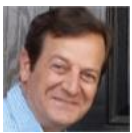
David Noever has research experience with NASA and the Department of Defense in machine learning and data mining. He received his Ph.D. from Oxford University, as a Rhodes Scholar, in theoretical physics.

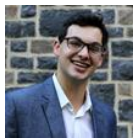
Matt Ciolino has experience in deep learning and computer vision. He received his bachelor's from Lehigh University in Mechanical Engineering.

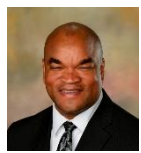
Gerry Dozier is a pioneer in artificial intelligence, machine learning, genetic algorithms, and evolutionary computation. He is a professor of computer science and software engineering and is committed to keeping our information secure. He holds a Charles D. McCrary Endowed Chair in the Samuel Ginn College of Engineering at Auburn University.